\documentclass[11pt]{article}

\usepackage[preprint]{acl}

\usepackage{times}
\usepackage{latexsym}

\usepackage[T1]{fontenc}

\usepackage[utf8]{inputenc}
\usepackage{xcolor}
\usepackage{colortbl}
\usepackage{multirow}

\usepackage{microtype}
\usepackage{makecell}
\usepackage{enumitem}

\usepackage{inconsolata}

\usepackage{graphicx}
\usepackage{booktabs}
\title{AfroXLMR-Social: Adapting Pre-trained Language Models \\ for African Languages Social Media Text}

\author{
\textbf{Tadesse Destaw Belay}$^{1,11}$, 
\textbf{Israel Abebe Azime}$^{2}$, 
\textbf{Ibrahim Said Ahmad}$^{3,4}$, \\
\textbf{David Ifeoluwa Adelani}$^{5,6}$, 
\textbf{Idris Abdulmumin}$^{7}$, 
\textbf{Abinew Ali Ayele}$^{8}$, \\
\textbf{Shamsuddeen Hassan Muhammad}$^{4,9}$, 
\textbf{Seid Muhie Yimam}$^{10}$ \\[1mm]
\footnotesize $^{1}$Instituto Politécnico Nacional, 
$^{2}$Saarland University,
$^{3}$Northeastern University,
$^{4}$Bayero University Kano, \\
\footnotesize $^{5}$Mila-Quebec AI Institute, McGill University,
$^{6}$Canada CIFAR AI Chair,
$^{7}$University of Pretoria,
$^{8}$Bahir Dar University, \\
\footnotesize 
$^{9}$Imperial College London,
$^{10}$University of Hamburg,
$^{11}$Wollo University\\
\footnotesize \texttt{Contact: tadesseit@gmail.com}
  }
  
\begin{document}
\maketitle
\begin{abstract}
Language models built from various sources are the foundation of today’s NLP progress. 
However, for many low-resource languages, the diversity of domains is often limited, more biased to a religious domain, which impacts their performance when evaluated on distant and rapidly evolving domains such as social media. Domain adaptive pre-training (DAPT) and task-adaptive pre-training (TAPT) are popular techniques to reduce this bias through continual pre-training for BERT-based models, but they have not been explored for African multilingual encoders. 
In this paper, we explore DAPT and TAPT continual pre-training approaches for African languages social media domain. We introduce \textbf{AfriSocial}, a large-scale social media and news domain corpus for continual pre-training on several African languages.  
Leveraging AfriSocial, we show that DAPT consistently improves performance (from 1\% to 30\% F1 score) on three subjective tasks: sentiment analysis, multi-label emotion, and hate speech classification, covering 19 languages. %
Similarly, leveraging TAPT on the data from one task enhances performance on other related tasks. For example, training with unlabeled sentiment data (source) for a fine-grained emotion classification task (target) improves the baseline results by an F1 score ranging from 0.55\% to 15.11\%. Combining these two methods (i.e. DAPT + TAPT) further improves the overall performance. The data and model resources are available at HuggingFace\footnote{\url{https://huggingface.co/tadesse/AfroXLMR-Social}}.
\end{abstract}

\section{Introduction}
Pre-trained language models (PLMs) are initially trained on vast and diverse corpora, including encyclopedias and web content \cite{conneau-2020-unsupervised,chiang-etal-2022-recent}. Subsequently, these pre-trained models are used in supervised training for a specific Natural Language Processing (NLP) task by further finetuning. Fine-tuned PLMs achieve strong performance across many tasks and datasets from various sources \cite{wu2024continuallearning}. However, this raises a question: Do PLMs function universally across domains, or does continual training of PLMs with domain-specific data offer better performance?

\begin{figure}[!t]
    \centering
    \includegraphics[width=\linewidth]{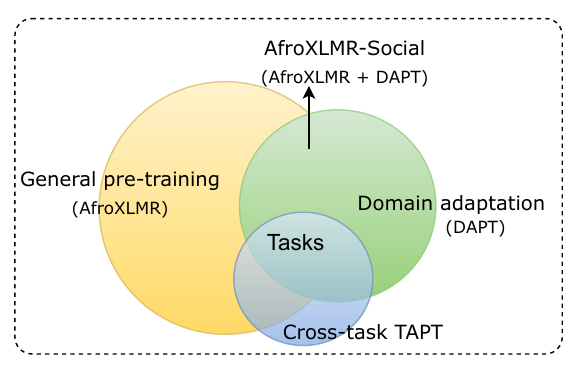}
    \caption{Continual pre-training illustration. A general-purpose pretrained model, such as AfroXLMR, is first adapted to a social domain, resulting in \mbox{AfroXLMR-Social}. This model then undergoes Cross-task TAPT using sentiment analysis, emotion, and hate speech data without the labels for further fine-tuning.}
    \label{fig:continual}
\end{figure}

While some studies have shown the benefit of continual pre-training on a domain-specific unlabeled data \cite{Lee_2019,gururangan-etal-2020-dont}, one domain may not be generalizable to other domains and languages. Moreover, it is unknown how the benefit of continual pre-training may vary with factors like the amount of unlabeled corpus, the source domain itself, the evaluation task, the resource richness of the target languages, and the trained target model \cite{gururangan-etal-2020-dont}. This raises the question of whether pre-training on a corpus more directly tied to the task can further improve performance. This work addresses these questions by continual domain adaptive pre-training (DAPT) and task adaptive pre-training (TAPT) on the downstream subjective NLP tasks in a low-resource language setup. We consider the social media domain (X) and News for a continual pre-training from a high-performing multilingual baseline model, AfroXLMR \cite{alabi-etal-2022-adapting}. AfroXLMR is a continually pre-trained model for African languages based on XLM-RoBERTa \cite{conneau-2020-unsupervised}. We explore subjective NLP task results from baseline (the base model result), DAPT, TAPT, and DAPT + TAPT, on a smaller but directly domain- and task-relevant unlabeled corpus. The results show that DAPT and TAPT highly benefit from similar source NLP tasks. Figure \ref{fig:continual} illustrates the general high-level continual pre-training strategies.
In summary, our contributions are:
\begin{itemize}[noitemsep]
    \item We present \textbf{AfriSocial}, a new quality domain-specific corpus for 14 African languages, collected from the social domain (X) and news.
    \item We perform a through analysis of domain and task adaptive continual pre-training across subjective NLP tasks for low-resource languages. 
    \item We achieve state-of-the-art results in the evaluated NLP tasks and publicly making \textbf{\mbox{AfriSocial}} social-domain corpus and \textbf{\mbox{AfroXLMR-Social}} pretrained models for developing low-resource languages. 
\end{itemize}


\begin{table*}[hbt!]
\centering
\resizebox{\textwidth}{!}{
\begin{tabular}{llclc}
\toprule
\textbf{Dataset name} &\textbf{Task name}&\textbf{\# lang.}&\textbf{Data Sources}& \textbf{\# Classes}\\ 
\midrule
AfriSenti  &Sentiment analysis&14&News, social media & 3 \\
AfriEmo  &Emotion analysis&17 &News, social media & 6\\
AfriHate  &Hate speech detection &15&News, social media & 3\\
\hline
\textbf{AfriSocial}  (new domain specific corpus) &Unlabeled&14&X and news & -\\
\bottomrule
\end{tabular}}
\caption{List of subjective NLP task evaluation datasets. Social media sources include posts/comments from YouTube and X. The AfriSenti dataset class labels are positive, negative, and neutral. The AfriEmo dataset labels are six basic emotions (anger, disgust, fear, joy, sadness, and surprise) in a multi-label annotation (an instance may have none, one, two, some, or all targeted emotion labels). AfriHate labels are abuse, hate, and neutral.}
\label{tab:source}
\end{table*}



\section{Related Work}
\noindent \textbf{Domain in NLP} Language models (LMs) pretrained on text from various sources are the foundation of today’s NLP. Domain adaptation in NLP refers to enhancing the performance of a model using similar domain data (target domain) by leveraging knowledge from an existing domain (source domain) \cite{ramponi-plank-2020-neural}. Domain refers to different implicit clusters of text representations in pretrained LMs, such as news articles, social media posts, medical texts, or legal documents \cite{aharoni2020domain,wu2024continuallearning}. Each domain has its unique characteristics, vocabulary, and writing style, which can affect the performance of NLP models when applied to new or unseen domains. Therefore, a similar domain means the text source from which the pre-trained model was made, similar to the target NLP task data source.

~\\ \textbf{Continual Pre-training of LMs}
There are several techniques for the downstream NLP task improvements, such as general-purpose pre-training \cite{wang-etal-2023-nlnde}, language-adaptive continual pre-training \cite{alabi-etal-2022-adapting}, domain-adaptive pre-training \cite{gururangan-etal-2020-dont}, task-adaptive pre-training \cite{alabi-etal-2022-adapting}, and data augmentation \cite{zhang-etal-2024-aadam}. Prior works have shown the benefit of continual pre-training using a domain-specific corpus \cite{gururangan-etal-2020-dont} and training LMs in a specific domain from scratch \cite{huang2020clinicalber}. However, continual pre-training is arguably more cost-effective than training from scratch, since it is a continuous pre-training from the existing base language model.

~\\ \textbf{Domain Adaptive Pre-training (DAPT)}
Domain Adaptive pre-training (DAPT) is straightforward, continuing pre-training a model on a corpus of unlabeled domain-specific text \cite{aharoni2020domain}.
DAPT techniques handle discrepancies between the source (pre-training) and target (fine-tuning) domains. Traditional fine-tuning often yields suboptimal results when pre-trained models encounter data that diverges significantly from their training data. In this regard, DAPT techniques reduce this mismatch by aligning data distributions, ensuring the model can generalize better in a better setup. \citet{Lee_2019} considers a single domain at a time and uses a language model pretrained on a smaller and less diverse corpus than the most varied and multilingual language models.

~\\ \textbf{Task Adaptive Pre-training (TAPT)}
Task-adaptive pre-training (TAPT) refers to pre-training on the unlabeled training set for a similar task-specific data \cite{gu-etal-2024-nlopt}. Compared to DAPT, TAPT solely leverages the training data of the similar downstream task for continuous pre-training. It uses a far smaller pre-training corpus that is much more task-relevant (assuming that the training set represents aspects of the task well). This makes TAPT much less expensive to run than DAPT. There are also a combination of various techniques, such as DAPT followed by TAPT, which is beneficial for end-task performance \cite{gururangan-etal-2020-dont}.

~\\ \textbf{Language Adaptive Pre-training (LAPT)} LAPT is also called language-adaptive fine-tuning (LAFT) \cite{alabi-etal-2022-adapting}. It focuses on adapting a pre-trained language model to a specific language(s) using any language-specific corpus. This is done by collecting any corpus for fine-tuning language models without considering the domain \cite{yu-joty-2021-effective,alabi-etal-2022-adapting}. LAPT is a vital step to improve language understanding and representation, especially for low-resource languages \cite{wang-etal-2023-nlnde}.  However, in the primary studies that investigated LAPT with various language-specific corpora, the impacts of the text sources on specific NLP tasks are unexplored. 

Although XLMR \cite{conneau-2020-unsupervised} pre-training corpus is derived from multiple sources and languages, it has not yet been explored whether these sources are diverse enough to generalize in a specific domain and task. This leads to asking whether subjective tasks related to social media text can be understood with this generic model. Towards this end, we explore further a continual training of DAPT and TAPT from AfroXLMR-\{76L\} \cite{adelani-etal-2024-sib}~\footnote{While the original AfroXLMR~\cite{alabi-etal-2022-adapting} cover 20 languages, we make use of the version with 76 languages~\cite{adelani-etal-2024-sib} with similar adaptation. Throughout this paper, AfroXLMR-76L is referred to as AfroXLMR.} and evaluate the impacts on highly subjective NLP tasks such as sentiment analysis, multi-label emotion, and hate speech classification.

\section{AfriSocial: Domain Adaptation Corpus}\label{sec:afrisocial}

\paragraph{Social Domain in our Work}
In computational linguistics, domain boundaries can be defined through various dimensions, including content, style, and purpose \cite{plank2016non}. In this work, we define the social domain based on two key factors: 1) Convergent textual characteristics, X and news sources provide a public discourse, comments, reactions, and conversational linguistic patterns that facilitate social interactions, and cultural expressions and 2) Functional similarity in downstream tasks, the selected evaluation task datasets are sourced from the two sources, shown in Table \ref{tab:source}. This grouping demonstrates comparable performance patterns when evaluating entity recognition models across X and news data, suggesting underlying linguistic commonalities despite surface differences and prioritizing functional and distributional similarities over source platform distinctions \cite{derczynski-etal-2016,ruder-plank-2018}.

We create AfriSocial, a social domain-specific corpus comprising X and news for 14 African languages, shown in Table \ref{tab:corpus}. 
We select X and news because they are the most common text sources for low-resource languages to annotate a dataset for supervised NLP tasks. The motivations behind creating this domain-specific corpus are the following.

\paragraph{Limited coverage for African languages} The available well-known compiled corpora are limited to include African languages; most of them are only English-centric, such as \texttt{fineweb} \cite{fineweb2-2024} and \texttt{C4} \cite{raffel2020exploring}. The reasons include the extra effort required to collect data for such low-resource languages, the limited availability of text, and the challenges in detecting the language and filtering sources.

\paragraph{Text quality issues} The quality of the available corpus is under consideration, especially for low-resource languages. For example, in the \texttt{OPUS} corpora \cite{opensubtitles2016}, there are Tigrinya (tir) texts under the Amharic (amh) file as both languages use the same script. Some of the available corpora are translated, such as OPUS-100 \cite{zhang-etal-2020-improving}, with the problem that the quality of the translator tool is still not mature enough for low-resource languages.

\paragraph{Non availability of social domain corpus} To be specific, a domain-specific corpus is vital for adapting pre-trained language models into a specific domain, such as a health-specific domain. Likewise, the social media corpus is limited even to high-resource languages.

\subsection{Data Sources Selection}
Based on our assessment, most of the NLP datasets of African languages are sourced and annotated from X and news domain, as sufficient text can be found in these two sources. As shown in Table \ref{tab:source}, subjective NLP tasks for African languages, such as sentiment analysis, multi-label emotion, and hate speech classification datasets, are sourced from X and news. The AfriSocial corpus is sourced from similar domains to further enhance these subjective tasks.  There is an X domain corpus and model for high-resource languages such as XLM-T \cite{XLMR-T:2022:LREC} to evaluate and improve task datasets sourced from X. However, no available corpora specialize in the social domain for low-resource African languages. More details about the data collections are presented in Appendix \ref{app:source}. 

\begin{table}[t]
\centering
\begin{tabular}{lrrr}
\midrule
\textbf{Lang.} &\textbf{X}& \textbf{News}&\textbf{Total Sent.}\\ 
\midrule
\texttt{amh}& 588,154 &45,480 & 633,634 \\
\texttt{ary}&9,219 & 156,494&165,712 \\
\texttt{hau}& 640,737&30,935 &671,672  \\
\texttt{ibo}&15,436 &38,231 &53,667  \\
\texttt{kin}&16,928 &72,583 &89,511  \\
\texttt{orm}&33,587 &59,429 & 93,016\\
\texttt{pcm}&106,577 &7,781 &116,358  \\
\texttt{som}&144,862 &24,473 &169,335  \\
\texttt{swa} & 46,588&--- &46,834  \\
\texttt{tir}&167,139 &45,033 &212,172  \\
\texttt{twi}& 8,681&--- & 8,681 \\
\texttt{yor}&26,560 & 49,591& 76,151 \\
\texttt{xho}& ---&354,959 & 354,959 \\
\texttt{zul}& 12,102&854,587 & 866,689 \\
\hline
\textbf{Total}&1.82M &1.74M & 3.56M\\
\bottomrule
\end{tabular}
\caption{AfriSocial corpus statistics at language and source level, where \textbf{Total Sent.} is the number of sentences. The full names of the languages are presented in Appendix \ref{app:lang}.}
\label{tab:corpus}
\end{table}

\subsection{Pre-processing and Quality Control}
We apply the following quality measures on the AfriSocial corpus. 
\paragraph{ Language Identification (LID)} We apply LID tools for each language. For example, for the language mixing problem of the existing corpus mentioned in Sec  §\ref{sec:afrisocial}, we used language-specific LID tools, GeezSwitch\footnote{\url{https://pypi.org/project/geezswitch/}} to handle Ethiopic script languages (Amharic and Tigrinya) and pycld3\footnote{\url{https://pypi.org/project/pycld3/}} for the supported Latin and other script languages at the sentence level. 

\paragraph{Sentence Segmentation} The same approach as language identification, we used tools for each language to segment into sentences. For the Ethiopic script language, we used \texttt{amseg} tool \cite{yimam-2021}, and for other Latin script languages, we used NLTK \cite{bird-loper-2004-nltk}. 

\paragraph{Other Preprocessing} We exclude sentences that contain hate/offensive words, very short sentences, only URL lines, and anonymize personally identifiable information (PII) such as usernames starting with the @ symbol, and email addresses. We pay special attention to ensuring that the available evaluation task data (sentiment, emotion, and hate speech) do not appear in the AfriSocial corpus before and after processing.  De-duplication is applied if a near-similar instance is present, excluding it from AfriSocial, not from the annotated dataset. Table \ref{tab:corpus} shows the AfriSocial corpus statistics with their sources and number of sentences. We did not perform further processing on the code-switching text as we trained one single multilingual model (AfroXLMR-Social), and we need code-switching or dialectal diversity to be captured in the model.

\begin{table*}[!htb]
\centering
\scalebox{0.85}{
\begin{tabular}{lrr|lrr|lrr}
\toprule
\multicolumn{3}{c|}{\textbf{AfriSenti}} & \multicolumn{3}{c|}{\textbf{AfriEmo}} & \multicolumn{3}{c}{\textbf{AfriHate}} \\ 
\cmidrule{2-3}\cmidrule{4-6}\cmidrule{7-9}
\textbf{Language } & \textbf{AfroXLMR} & \textbf{+DAPT} &\textbf{Language } & \textbf{AfroXLMR} & \textbf{+DAPT} & \textbf{Language } & \textbf{AfroXLMR} & \textbf{+DAPT}\\
\midrule
\texttt{amh} & 50.09 & \textbf{57.22} &\texttt{afr}&43.66&\textbf{44.57}&\texttt{amh}&73.54&\textbf{78.57}\\
\texttt{arq} & 52.22 & \textbf{64.62} &\texttt{amh}&68.97&\textbf{71.67}&\texttt{arq}&43.41&\textbf{45.96}\\
\texttt{ary} & 52.86 & \textbf{62.34} &\texttt{ary}&47.62 &\textbf{52.63}&\texttt{ary}&75.13&\textbf{75.6}\\
\texttt{hau} & 79.34 & \textbf{81.66} &\texttt{hau}&64.30 &\textbf{70.74}&\texttt{hau}&\textbf{81.55}&80.78\\
\texttt{ibo} & 76.92 & \textbf{79.8} &\texttt{ibo}&26.27 &\textbf{54.54} &\texttt{ibo}&82.78&\textbf{88.05}\\
\texttt{kin} & 70.95 & \textbf{72.73} &\texttt{kin}&52.39 &\textbf{56.73}&\texttt{kin}&75.28&\textbf{78.75}\\
\texttt{pcm} & 50.47 & \textbf{52.09} &\texttt{orm}&52.28 &\textbf{61.38}&\texttt{orm}&67.23&\textbf{74.11}\\
\texttt{por} & 60.93 & \textbf{64.81} &\texttt{pcm}&55.39 &\textbf{59.93}&\texttt{pcm}&64.85&\textbf{67.61}\\
\texttt{swa} & 28.26 & \textbf{61.42}&\texttt{ptMZ}&22.09 &\textbf{36.80}&\texttt{som}&\textbf{55.66}&55.64\\
\texttt{tso} & 35.37 & \textbf{38.81} &\texttt{som}&48.78 &\textbf{54.86} &\texttt{swa}&91.51&\textbf{91.2}\\
\texttt{twi} & 47.2 & \textbf{56.00} &\texttt{swa}&30.74 &\textbf{34.35}&\texttt{tir}&50.2&\textbf{55.9}\\
\texttt{yor} & 72.27 & \textbf{74.63}&\texttt{tir}&57.22 &\textbf{60.71}&\texttt{twi}&46.89&\textbf{48.42}\\
\texttt{orm}&20.09 &\textbf{24.28}&\texttt{vmw}&21.18 &\textbf{22.08} &\texttt{xho}& 50.91 & \textbf{59.17}\\
\texttt{tir}&22.45 &\textbf{24.53}&\texttt{yor}&28.65 &\textbf{39.26} &\texttt{yor} & 53.44 & \textbf{77.9}\\
\midrule
\textbf{Avg.}&51.39 &\textbf{58.21}&\textbf{Avg.}&44.25 &\textbf{51.45} &\textbf{Avg.} & 65.17 & \textbf{69.83}\\

\bottomrule
    \end{tabular}
    }
    \caption{Result of baseline (AfroXLMR) and DAPT (AfroXLMR-Social) across the three datasets (AfriSenti, AfriEmo, and AfriHate). During TAPT, the text for the task-adaptive data is without the labels, and the evaluation is cross-tasked among the three target datasets. Reported results are macro-F1.}
    \label{tab:dapt}
\end{table*}

\section{Evaluation Tasks and Datasets} \label{eva-data}
We select subjective NLP tasks for our evaluation based on the following reasons. 1) Subjective tasks face more disagreement during annotations, leading to less performance in the evaluation, especially for low-resource languages \cite{fleisig-etal-2023-majority,belay-etal-2025-culemo}. As we can see from the SemEval-2025 Task 11 \cite{muhammad-etal-2025-semeval}, an emotion detection shared task covering 32 languages, low-resource languages are not well explored, and the lowest results are from African languages.
2) Subjective NLP tasks of African languages are sourced from X and news, as shown in Table \ref{tab:source}. These sources align with the same domain as the AfriSocial corpus. The three subjective tasks for our evaluation are sentiment analysis, multi-label emotion detection, and hate speech classification. We keep the original train-test split of all evaluation datasets throughout our experiment for proper comparison with the benchmark results.

\subsection{AfriSenti: Sentiment Analysis Dataset} 
AfriSenti \cite{muhammad-etal-2023-afrisenti} is a sentiment analysis dataset across 14 African languages. It aggragates some existing datasets such as NaijaSenti~\cite{muhammad-etal-2022-naijasenti}, Amharic Twitter sentiment~\cite{yimam-etal-2020-exploring}, and manually curated data.  The data is sourced from X (formerly Twitter) and annotated in one of the three sentiment classes: positive, negative, and neutral. From 14 languages, the two languages, Oromo (\texttt{orm}) and Tigrinya (\texttt{tir}) have only test sets.


\subsection{AfriEmo: Multi-label Emotion Dataset}
SemEval-2025-Task 11 \cite{muhammad-etal-2025-semeval} is an emotion dataset that covers 32 languages, from diverse domains such as social media platforms (X, Reddit, YouTube, and others) and news. The AfriSocial domain-specific corpus targets low-resource African languages; we target the African languages emotion dataset from the SemEval-2025 Task 11, specifically from BRIGHTER \cite{muhammad-etal-2025-brighter} and EthioEmo \cite{belay2025enhancing}, which we call \textbf{AfriEmo}. It covers 17 African languages from the 32 languages. This dataset is annotated in a multi-label approach - a text might have any combination (none, one, some, or all) of emotion labels from a given set of emotions (anger, disgust, fear, joy, sadness, and surprise).

\subsection{AfriHate: Hate Speech Dataset}
AfriHate \cite{muhammad2025afrihate} is a multilingual hate and abusive speech dataset in 15 African languages sourced from X. Each text is categorized into one of the abusive, hate, or neutral labels. The languages covered in the corresponding evaluation datasets, such as language name, ISO code, countries/regions spoken, language family, and writing script. See the details in Appendices \ref{app:lang} and \ref{app:datastat}.

\begin{table*}[!htb]
\centering
\resizebox{\textwidth}{!}{

\begin{tabular}{lcccc|cccc|cccc}
\toprule
 &\multicolumn{4}{c|}{\textbf{AfriSenti TAPT}} & \multicolumn{4}{c|}{\textbf{AfriEmo  TAPT }} & \multicolumn{4}{c}{\textbf{AfriHate TAPT }} \\ 
\cmidrule{2-5}\cmidrule{6-9}\cmidrule{10-13}
    \textbf{Lang.}& \textbf{Base} & \textbf{AfriEmo} & \textbf{AfriHate} & \textbf{\makecell{DATP\\+ TAPT}} & \textbf{Base} & \textbf{AfriSenti} & \textbf{AfriHate} & \textbf{\makecell{DATP\\+ TAPT}} & \textbf{Base} & \textbf{AfriSenti} & \textbf{AfriEmo} & \textbf{\makecell{DATP\\+ TAPT}}\\
\midrule
\texttt{amh} & 50.09 & 54.91 & 54.48 & \textbf{55.80} & 68.97 & 69.56 & 67.21 & \textbf{71.04} & 73.54 & 73.13 & 73.26 & \textbf{78.06}\\
\texttt{arq}  & 52.22 & 62.42 & 59.41 & \textbf{63.38} & 44.93 &51.25 & 48.95  & 48.72   & 43.41 & \textbf{46.19} & 43.84 & 44.21\\
\texttt{ary} & 52.86 & \textbf{64.13} & 52.80 & 63.05  & 47.62 & 50.21 & 48.81 & \textbf{51.63} & 75.13 & 75.13 & 71.70 & \textbf{77.51} \\
\texttt{hau} & 79.34 & 80.65 & 80.08 & \textbf{82.71} & 64.30 & 66.93 & 61.16 & \textbf{69.77}  & 81.55 & 77.76 & 81.94 & \textbf{82.09} \\
\texttt{ibo} & 76.92 & 80.01 & 78.29 & \textbf{80.42} & 26.27 & 53.13 & 51.26 & \textbf{54.26}  & 82.78 & \textbf{87.69} & 86.52 & \textbf{87.68}\\
\texttt{kin} & 70.95 & \textbf{71.47} & 69.72 & 69.53 & 52.39 & 53.27 & 53.49 & \textbf{54.47}  & 75.28 & 77.48 & 76.30 & \textbf{78.36}\\
\texttt{orm} & 20.09 & 23.53 & \textbf{42.99} & 28.93 & 52.28 & 56.43 & 52.22 & \textbf{58.75} & \textbf{77.08} & 69.03 & 67.67 & 71.07 \\
\texttt{pcm} & 50.47 & 50.97 & 50.29 & \textbf{52.04} & 55.39 & 56.71 & 56.60 & \textbf{58.89} & 64.85 & 67.73 & 66.94 & \textbf{69.91}\\
\texttt{ptMZ} & 60.93 & \textbf{64.05} & 62.80 & 63.75 & 22.09 & 37.20 & 30.99 & \textbf{37.76} &  ---     & ---     & ---    & --- \\
\texttt{som}  &  ---     & ---     & ---    & ---  & 48.78 & 50.33 & 49.63 & \textbf{52.65}  & 55.66 & \textbf{57.29} & 54.97 & 56.75 \\
\texttt{swa}  & 28.26 & \textbf{59.33} & 57.26 & 54.94 & 30.74 & \textbf{33.02} & 31.85 & 32.74 & 91.51 & \textbf{91.91} & 91.27 & 91.16\\
\texttt{tir}  & 22.45 & 10.81 & 16.22 & \textbf{28.90} & \textbf{57.22 }& 55.72 & 55.84 & 57.12 & 50.20 & 54.21 & \textbf{56.98} & 32.70\\
\texttt{twi}  & 47.20 & 47.68 & 50.23 & \textbf{54.47} &  ---     & ---     & ---    & ---   & 46.89 & \textbf{49.30} & 48.94 & 49.01 \\
\texttt{yor}  & 72.27 & 72.22 & 70.90 & \textbf{73.65} & 28.65 & 34.34 & 32.87 & \textbf{35.89} & 53.44 & 53.69 & 54.51 & \textbf{54.76}\\
\midrule
\textbf{Avg.} & 52.62 & 57.09 & 57.34 & \textbf{59.35} & 45.74 & 51.39 & 49.22 & \textbf{52.87} & 66.72 & 67.46 & 67.14 & \textbf{67.77} \\

\bottomrule
    \end{tabular}}
    \caption{Cross-TAPT results across the three datasets. \textbf{Base} column is baseline results from AfroXLMR, DATPT + TAPT column results for AfriSenti and AfriHate are TAPT from the AfriEmo dataset.  DATPT + TAPT for AfriEmo results is TAPT from the AfriSenti dataset. Reported results are the Macro F1 score. Blank values (---) indicate that the specific dataset does not cover the language.}
    \label{tab:tapt}
\end{table*}

\section{Language Models}
\subsection{Encoder-only Language Models}
Multilingual encoder-only pretrained language models (PLMs) such as XLM-R \cite{conneau-2020-unsupervised} and mBERT  \cite{devlin2019bert}  have shown impressive capability on many languages for a variety of downstream tasks. They are also often used to initialize checkpoints to adapt to other languages, such as AfroXLMR-76L \cite{adelani-etal-2024-sib}, which is initialized from XLM-R to specialize in African languages. 
We evaluate popular multilingual and African-centric PLMs such as AfroLM \cite{dossou-etal-2022-afrolm} and AfriBERTa \cite{ogueji-etal-2021-small} and found that AfroXLMR is better for the targeted evaluation datasets as it covers more African languages. We make AfroXLMR our benchmarks and further train on the AfriSocial domain and the selected evaluation tasks. The AfroXLMR followed by DAPT with the AfriSocial domain-specific corpus gives us \textbf{\mbox{AfroXLMR-Social}}.
The detailed hyperparameters of the continual training are shown in Appendix \ref{app:params}.


\subsection{Large Language Models}
We compare our DAPT and TAPT approach results from AfroXLMR with state-of-the-art open source LLMs such as Llama 3 \cite{grattafiori2024llama}, Gemma 2 \cite{team2024gemma2}, Mistral \cite{jiang2024mixtral}, and proprietary LLMs such as Gemini 1.5 \cite{team2024gemini} and GPT-4o \cite{openai2024gpt}.  For AfriSenti and AfriHate task results, we use the LLMs benchmark results from the \textbf{AfroBench} \cite{afrobench} leaderboard. For the AfriEmo task, we use LLM results from the SemEval-2025 Task 11 datasets papers \cite{muhammad-etal-2025-brighter,belay-etal-2025-evaluating}. The detailed versions of the LLMs are presented in the Appendix \ref{app:llms-details}.

\begin{figure*}[!t]
    \centering
    \includegraphics[width=\linewidth]{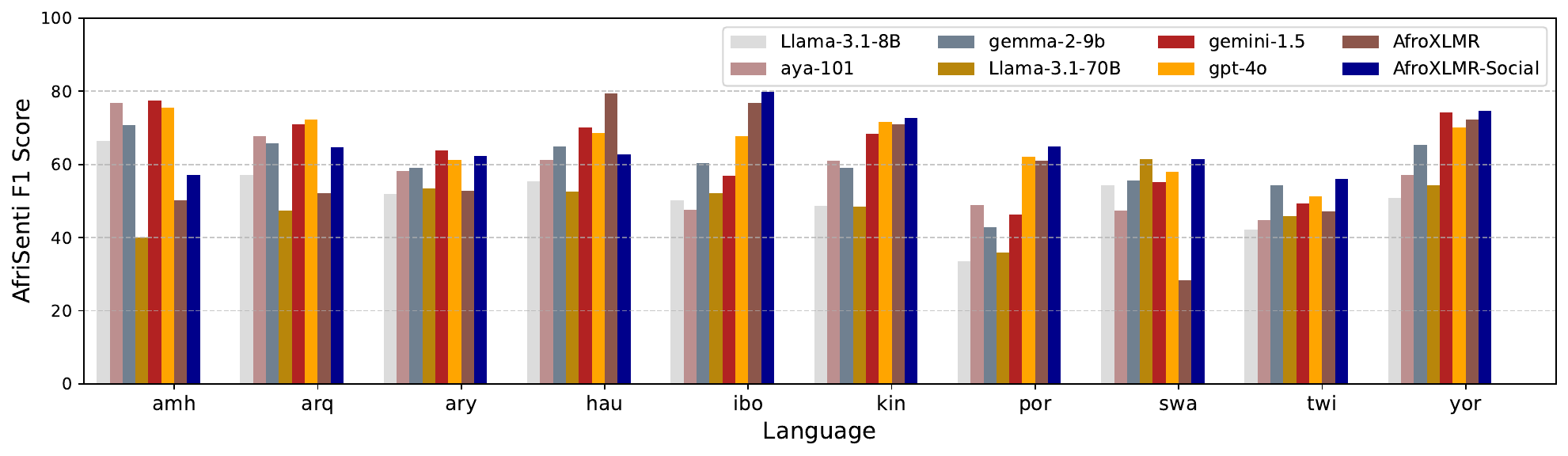}
    \caption{\textbf{AfriSenti} Macro F1 results from AfroXLMR-Social and LLMs. The results for LLMs are based on \textbf{zero-shot} evaluations, selecting the best results from five different types of prompts. The benchmark results for the LLMs are taken from the AfroBench \cite{afrobench} leaderboard.}
    \label{fig:afrisenti-llms}
\end{figure*}

\begin{figure*}[!t]
    \centering
    \includegraphics[width=\linewidth]{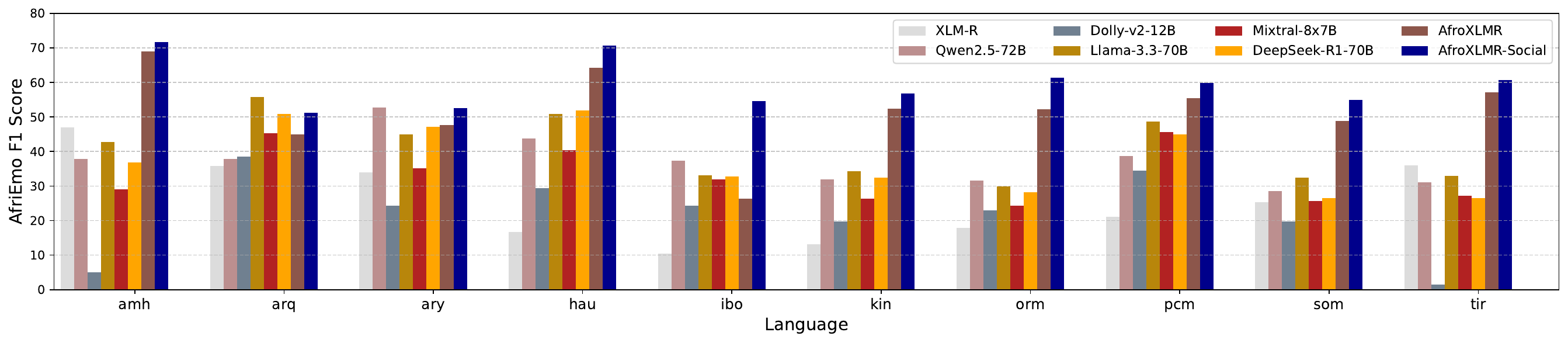}
    \caption{\textbf{AfriEmo} Macro F1 results from AfroXLMR-Social and LLMs. The results for LLMs are based on \textbf{zero-shot} evaluations, selecting the best results from five different types of prompts. The benchmark results for the LLMs are taken from the SemEval-2025 Task 11 dataset papers \cite{muhammad-etal-2025-brighter,belay2025enhancing}.}
    \label{fig:afriemo-llms}
\end{figure*}

\begin{figure*}[!t]
    \centering
    \includegraphics[width=\linewidth]{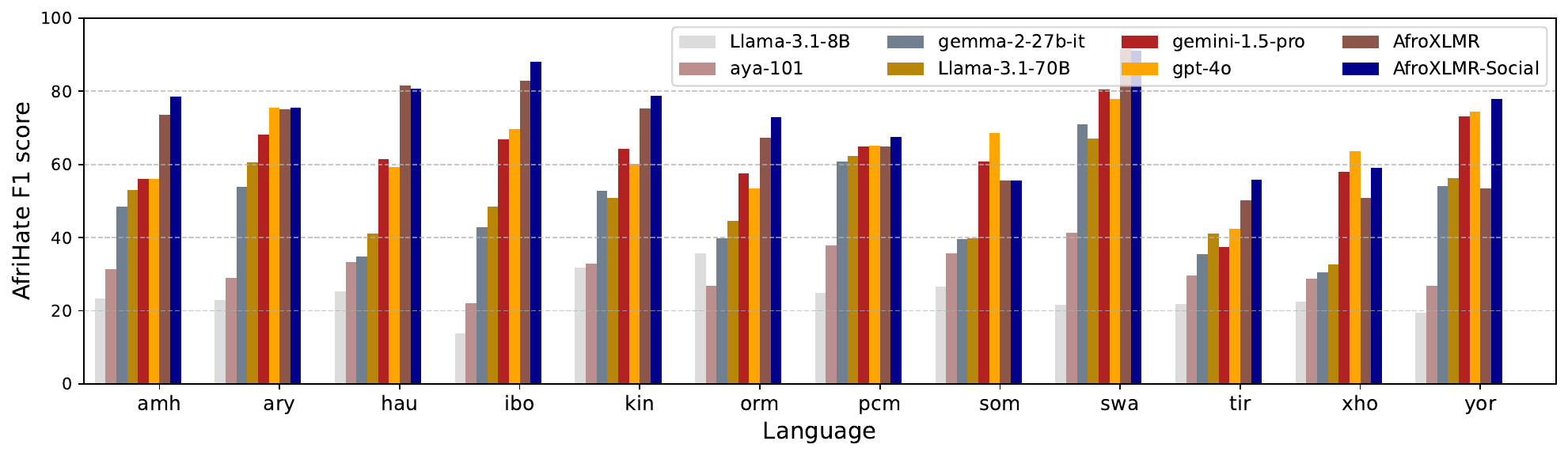}
    \caption{\textbf{AfriHate} Macro F1 results from AfroXLMR-Social and LLMs. The results for LLMs are based on \textbf{zero-shot} evaluations, selecting the best results from five different types of prompts.. The baseline results for the LLMs are taken from the AfroBench \cite{afrobench} leaderboard.}
    \label{fig:afrihate-llms}
\end{figure*}

\section{Experiment Results}
\subsection{Domain Adaptive Pre-training (DAPT)} \label{sec:dapt}
The domain-adaptive pre-training (DAPT) approach is straightforward; we continue pre-training from the strong baseline language model (AfroXLMR) using the domain-specific AfriSocial corpus in a multilingual setup. Baseline results from (AfroXLMR) and after applying DAPT are presented in Table \ref{tab:dapt}. 

~\\ \textbf{Baseline}: As a baseline, we evaluate various encoder-only models and found that AfroXLMR, which covers 76 African languages \cite{alabi-etal-2022-adapting}, performs better than other BERT-based encoder-only models across targeted datasets since it includes more African languages during pre-training. The evaluation results of othe encoder-only multilingual and African-centric models are shown in Appendix \ref{app:baseline}.

~\\ \textbf{Results}: The results before and after DAPT are shown in  Table \ref{tab:dapt}, AfroXLMR and DAPT columns, respectively. We observe that DAPT improves over the baseline in almost all languages and datasets, demonstrating the benefit of DAPT when the target domain is relevant. It consistently improves over the baseline models for each language. It suggests that continual pre-training on a small, quality, domain-relevant dataset is important for subjective tasks from the same domain.

\subsection{Task Adaptive Pre-training (TAPT)}
Similar to DAPT, TAPT consists of a second phase of continual pre-training. TAPT is a cross-task transfer across datasets, which refers to finetuning on the unlabeled data of the non-targeted task during evaluation. We explore the TAPT approach by directly applying it to the base model followed by DAPT. For example, if we make TAPT for the AfriSenti task, we further fine-tune the base model and DAPT model using the unlabeled data of the AfriEmo and AfriHate datasets separately. Compared to DAPT, the task-adaptive approach strikes a different trade-off: it uses a far smaller pre-training corpus, assuming the training set represents aspects of the target task. This makes TAPT much less expensive to run than DAPT. When we apply TAPT across tasks, for example, TAPT with AfriSenti for AfriHate evaluation, we ensure that we exclude any duplication from the training data (in this case, AfriSenti) to prevent data leakage.

~\\ \textbf{Results} Table \ref{tab:tapt} shows the TAPT results across datasets and languages. As a result, all TAPT performs better than the base model and equal or competitive results with the DAPT. This indicates we can achieve better results for our targeted evaluation task using very small, quality, and task-related data. In our case, using AfriSenti data without the labels as a TAPT is very helpful for the fine-grained multi-label emotion classification task. For the AfriSenti sentiment evaluation task, TAPT with AfriHate data achieves a better average results than TAPT with AfriEmo. For the AfriHate evaluation, TAPT in AfriSenti data has better average results than TAPT with AfriEmo. In addition to the task similarity, this improvement might be affected by the number of total instances in each dataset (the total instances of AfriSenti is 107,694, AfriEmo 70,859, and AfriHate 90,455) and the vocabulary similarity across the datasets. 

\subsection{Combining DAPT and TAPT}
We explore the effect of both adaptation techniques by combining DAPT and TAPT. In this approach, we apply DAPT to the base model and then the TAPT training. These phases of pre-training add up to make this approach the more computationally expensive setting. 

~\\ \textbf{Results} The results are shown in Table \ref{tab:tapt}, DAPT + TAPT column. The results show that the subjective NLP tasks benefit from the combined DAPT and TAPT approaches. DAPT followed by TAPT achieves the best performance.  However, first adapting the model to the domain (DAPT), then applying TAPT would be susceptible to catastrophic forgetting of the domain-relevant corpus \cite{yogatama2019learning}; alternate methods of combining the procedures may result in better downstream performance. This is shown from the summarized result in Table \ref{tab:summ} that sometimes the DAPT + TAPT result performs less than the DAPT-only results, while it is better than the baseline results.

\subsection{AfroXLMR-Social Vs. LLMs}
This section compares our AfroXLMR-Social results with state-of-the-art open-source and proprietary Large Language Models (LLMs). Table \ref{tab:summ2} shows the summarized average results across tasks. AfroXLMR-Social leads the performance over LLMs across the three targeted tasks. 
Figures \ref{fig:afrisenti-llms}, \ref{fig:afriemo-llms}, and \ref{fig:afrihate-llms} present the comparison results for the base model (AfroXLMR), AfroXLMR-Social, and both open-source and proprietary LLMs across the AfriSenti, AfriEmo, and AfriHate datasets, respectively. The result indicates that the domain-specialized encoder-only model has better or comparable results with LLMs. Generally, it means that we still need encoder-only models for low-resourced African languages and suggests that future LLMs are expected to include more training data for underrepresented languages.

\begin{table}[h!]
\centering
\resizebox{\columnwidth}{!}{
\begin{tabular}{llc}
\hline
 \textbf{Dataset}&\textbf{Models} & \textbf{Avg.} \\
\hline
\multirow{6}{*}{{\textbf{AfriSenti}}} 
& AfroXLMR  & 51.39  \\
& \quad + DAPT (AfroXLMR-Social)  & 56.85  \\
&\quad + TAPT (AfriEmo)& 55.72  \\
&\quad + TAPT (AfriHate)& 55.83  \\
&\quad + DAPT + TAPT (AfriHate)& \underline{56.91}  \\
&\quad + DAPT + TAPT (AfriEmo)& \textbf{57.73}  \\
\hline
\multirow{6}{*}{{\textbf{AfriEmo}}} 
& AfroXLMR  & 44.30  \\
&\quad + DAPT (AfroXLMR-Social)   & \textbf{51.48}  \\
&\quad + TAPT (AfriSenti)& 49.14  \\
&\quad + TAPT (AfriHate)& 47.12  \\
&\quad + DAPT + TAPT (AfriSenti)& \underline{49.84}  \\
&\quad + DAPT + TAPT (AfriHate)& 48.93  \\
\hline
\multirow{6}{*}{{\textbf{AfriHate}}} 
& AfroXLMR  & 67.03  \\
&\quad + DAPT (AfroXLMR-Social)   & \textbf{70.56}  \\
&\quad + TAPT (AfriSenti)& 67.30  \\
&\quad + TAPT (AfriEmo)& 67.73  \\
&\quad + DAPT + TAPT (AfriSenti)& 66.55  \\
&\quad + DAPT + TAPT(AfriEmo)& \underline{67.18}  \\
\hline
\end{tabular}
}
\caption{Summary results of Table \ref{tab:tapt}, the average of the Macro F1 score across languages. \textbf{Boldface} values are the overall best scores for the specific dataset, and results with \underline{underlines} are the best cross-TAPT dataset.}
\label{tab:summ}
\end{table}

\begin{table}[!h]
\centering
\resizebox{\columnwidth}{!}{
\begin{tabular}{lcc|lc}
\midrule
\textbf{Model} &AfriSenti& AfriHate&\textbf{Model}&AfriEmo\\
\midrule
Gemma-1.1-7B& 39.7 &24.3 &LaBSE & 35.7 \\
Llama-2-7B  & 38.9 &21.9 &RemBERT & 26.8 \\
Llama-3-8B  & 41.8 &27.9 &XLM-R & 23.4 \\
LLaMAX3-8B  & 49.8 &28.6 &mBERT & 23.0 \\
Llama-3.1-8B& 41.8 &23.6 &mDeBERTa & 26.7 \\
gemma-2-9B  & 55.5 &29.9 &Qwen2.5-72B & 35.3 \\
Aya-101-13B & 57.1 &30 &Dolly-v2-12B & 21.1 \\
gemma-2-27B & 58.6 &45.5 &Mixtral-8x7B & 31.4 \\
Llama-3.1-70B&46.9 &49 &Llama-3.3-70B & 38.3 \\
Gemini-1.5 pro   & 62.6&62.1&DeepSeek-70B & 36.6 \\
GPT-4o      & 62.6 &63.5 &AfroXLMR & 44.3 \\
\midrule
{\small \textbf{AfroXLMR-Social}}&\textbf{57.7}&\textbf{68.8 }&{\small \textbf{AfroXLMR-Social}} & \textbf{51.5} \\
\bottomrule
\end{tabular}}
\caption{ Summary results on fine-tuned models and LLMs. The results show the average Macro F1 score across all languages in the corresponding datasets: AfriSenti - average of 14 languages, AfriHate - average of 15 languages, and AfriEmo is the average of 17 languages.}
\label{tab:summ2}
\end{table}

\section{Conclusion}
This work explored the effects of domain-adaptive pre-training (DAPT) and task-adaptive pre-training (TAPT) across three subjective tasks involving African languages.  We created the AfriSocial corpus, a social domain-specific corpus sourced from X and news. Using AfriSocial, we further developed the AfroXLMR-Social language model, which specialized in the social domain. We improved the performance of evaluated tasks, sentiment analysis (AfriSenti), emotion analysis (AfriEmo), and hate speech classification (AfriHate) using DAPT and TAPT techniques. We showed that pre-training the model towards a small domain-specific corpus and related task-relevant data can provide significant improvements. While the combination of the two methods, DAPT + TAPT, also achieved better results than the baseline models, TAPT followed by DAPT would be susceptible to catastrophic forgetting of the task-relevant corpus. We achieved better state-of-the-art results using a small domain-related corpus from the encoder-only model than state-of-the-art large-language models (LLMs). AfriSocial and AfroXLMR-Social will support the development of African languages in the NLP and improve similar-sourced tasks. It opened further domain explorations as the AfriSocial X and news domains are also available separately.  

\section*{Limitations}
Our work is not without limitations. We identify the following limitations with its future suggestions.
\paragraph{Domain Coverage.} This work focuses on social media data from X and news sources for downstream tasks that are inherently subjective, such as sentiment analysis, emotion recognition, and hate speech classification. Extending the evaluation to out-of-domain data (e.g., health forums, long-form blogs) and the impacts of topic variations (e.g., politics, sports, business, health) presents another promising avenue for understanding cross-domain generalization in social media–based tasks. For the domain-adaptive pre-training (DAPT) exploration, we utilized a corpus of 3.5M sentences, which exhibits substantial variation in data statistics across different languages.

\paragraph{Evaluation Tasks.} In this work, we restrict our evaluation to three subjective tasks—sentiment analysis, emotion recognition, and hate speech classification—in order to highlight the effects of DAPT and TAPT approaches within the social domain. Future work could extend these approaches to a broader range of downstream NLP tasks, including more knowledge-intensive and objective benchmarks such as question answering and machine translation, thereby offering a more comprehensive understanding of their generalizability and impact.

\paragraph{Evaluation of LLMs} Assessing the impact of DAPT and TAPT approaches on the latest LLM families—such as Llama, Gemini, GPT, Mistral, and others remains an open direction for future research. In-context learning evaluations, particularly in few-shot settings, provide a promising lens for understanding model behavior, while prompting strategies such as Chain-of-Thought (CoT) reasoning and in-domain prompting have demonstrated notable improvements in LLM performance across various tasks. Systematic evaluation of these techniques in combination with DAPT and TAPT may therefore yield more profound insights and potentially lead to different conclusions regarding the effectiveness and generalizability of such adaptation methods.

\paragraph{Imbalanced Data Across Languages.} As illustrated in the AfriSocial corpus (Table \ref{tab:corpus}), there exists substantial variability in the availability of domain-specific data across languages (e.g., 8.6K sentences for Twi versus 866K for Zulu). Investigating the impact of such imbalances on the effectiveness of DAPT and TAPT continual pre-training approaches could yield valuable insights into both the robustness of adaptation techniques and language-specific behaviors. Incorporating more balanced data across languages in future work may further enhance the evaluation and provide a clear understanding of the dynamics of cross-lingual adaptation.
\section*{Acknowledgments}
The work was done with partial support from the Mexican Government through the grant A1-S-47854 of CONACYT, Mexico, grants 20241816, 20241819,  and 20240951 of the Secretaría de Investigación y Posgrado of the Instituto Politécnico Nacional, Mexico. The authors thank the CONACYT for the computing resources brought to them through the Plataforma de Aprendizaje Profundo para Tecnologías del Lenguaje of the Laboratorio de Supercómputo of the INAOE, Mexico and acknowledge the support of Microsoft through the Microsoft Latin America PhD Award. 
\newline
Shamsuddeen acknowledges support from a Research Grant under the Nigeria Artificial Intelligence Research (NAIR) Scheme, administered by the National Information Technology Development Agency (NITDA), for developing a dataset for Nigerian languages.

\bibliography{custom}
\newpage
\onecolumn
\appendix

\section*{Appendix}

\section{Languages Covered in Evaluation}\label{app:lang}
Table \ref{tab:lang} shows the 19 language details we evaluated in this work across different tasks.

\begin{table*}[hbt!]
\centering
\resizebox{\textwidth}{!}{
\begin{tabular}{llllllc}
\midrule
\textbf{Language} & \textbf{ISO} &\textbf{Subregion}& \textbf{Spoken in}&\textbf{Lang. family} & \textbf{Script}& \textbf{\# Speakers}\\
\midrule
Afrikaans&\texttt{afr}&South Africa&South Africa, Namibia, Botswana &Indo-European&Latin &7M\\
Amharic         &\texttt{amh} &East Africa & Ethiopia, Eritrea &Afro-Asiatic  &Ethiopic&57M\\
Algerian Arabic  &\texttt{arq} &North Africa & Algeria &Afro-Asiatic    &Arabic&36M\\
Moroccan Arabic &\texttt{ary} &North Africa &Morocco  &Afro-Asiatic    &Arabic/Latin&29M\\
Hausa  &\texttt{hau} &West Africa &Northern Nigeria, Ghana, Cameroon &Afro-Asiatic&Latin&77M\\
Igbo   &\texttt{ibo} &West Africa &Southeastern Nigeria& Niger-Congo&Latin&31M\\
Kinyarwanda&\texttt{kin} &East Africa &Rwanda &Niger-Congo &Latin&10M\\
Oromo      &\texttt{orm} &East Africa &Ethiopia&Afro-Asiatic&Latin&37M\\
Nigerian Pidgin&\texttt{pcm} &West Africa &Nigeria, Ghana, Cameroon &English-Creole &Latin&121M\\
Mozambican Portug. &\texttt{ptMZ} &Southeastern Africa&Mozambique &Indo-European &Latin&13M\\
Somali     &\texttt{som} &East Africa &Ethiopia, Kenya, Somalia   &Afro-Asiatic &Latin&22M\\
Swahili    &\texttt{swa} &East Africa &Tanzania, Kenya, Mozambique&Niger-Congo  &Latin&>71M\\
Tigrinya   &\texttt{tir} &East Africa &Ethiopia, Eritrea  &Afro-Asiatic  &Ethiopic&9M\\
Twi   &\texttt{twi} &West Africa &Ghana  &Niger-Congo  &Latin&9M\\
Makhuwa    &\texttt{vmw} &East African&Mozambique, Tanzania &Niger–Congo&Latin&7M\\
Xitsonga &\texttt{tso} & Southern Africa &South Africa, Zimbabwe, Mozambique & Niger-Congo& Latin&7M\\
Xhosa      &\texttt{xho}&Southern Africa&South Africa, Zimbabwe, Lesotho&Niger-Congo&Latin&19M\\
Yoruba     &\texttt{yor} &West Africa &Southwestern, Central Nigeria, Togo  &Niger-Congo&Latin&46M\\
Zulu       &\texttt{zul} &Southern Africa &South Africa &Niger-Congo& Latin&29M\\
\bottomrule
\end{tabular}}
\caption{Additional information on the African languages evaluated in this work: ISO-3 digit language code, region spoken, the family of the language, its script, and number of L1 and L2
speakers.}
\label{tab:lang}
\end{table*}

\section{Evaluation Data statistics}\label{app:datastat}
Table \ref{tab:dataset-split} shows the train-test split of the evaluation datasets AfriSenti, AfriEmo, and AfriHate.

\begin{table*}[!htb]
    \centering
    \resizebox{\textwidth}{!}{
    \begin{tabular}{lcccc|rrrr|rrrr}
    \toprule
    \multirow{2}{*}{\textbf{Lang.}} &\multicolumn{4}{c|}{\textbf{AfriSenti}} & \multicolumn{4}{c|}{\textbf{AfriEmo}} & \multicolumn{4}{c}{\textbf{AfriHate}} \\ \cmidrule{2-5}\cmidrule{6-9}\cmidrule{10-13}
        & \textbf{Train} & \textbf{Dev} & \textbf{Test} & \textbf{Total} & \textbf{Train} & \textbf{Dev} & \textbf{Test} & \textbf{Total} & \textbf{Train} & \textbf{Dev} & \textbf{Test} & \textbf{Total}\\
    \midrule
afr  & -& - &- &-            & 2107  & 98   & 1065 & 3270  &- & -&- &- \\
amh  &5985& 1498&2000 &9483   & 3549  & 592  & 1774  & 5915 &3467 &744 &747 &4958 \\
arq  &1952 &415 &959 &3062    & 901  & 100   & 902  & 1903  &716 & 211&323 &1250 \\
ary  &5584 &1216 &2962 &9762  & 1608  & 267  & 812   & 2687 &3240 &695 &699 &4634 \\
hau  &14173 &2678 &5304 &22155&2145  &356  & 1080  & 3581  &4566 &1029 &1049 & 6644\\
ibo  &10193 &1842 &3683 &15718 & 2880  & 479  & 1444  & 4803 &3419 &774 &821 &5014 \\
kin  &3303 &828 &1027 &5158  & 2451  & 407  & 1231  & 4089 &3302 &706 &714 &4722 \\
orm  &--- & 397&2097 &2494     & 3442  & 575  & 1721  & 5738 &3517 &763 &759 &5039 \\
pcm  &5122 &1282 &4155 &10559& 3728  &620  & 1870  & 6218 &7416 &1590 &1593 &10599 \\
ptMZ &3064 &768 &3663 &7495  & 1546  & 257  & 776   & 2579 &--- &--- &--- & ---\\
som  &---  & --- & --- & --- & 3392  & 566  & 1696  & 5654 &3174 &741 &745 & 4660\\
swa  &1811 &454 &749 &3014 & 3307  & 551  & 1656  & 5514 &14760 &3164 & 3168&21092 \\
tir  &---  &399 &2001 &2400 & 3681  & 614  & 1840  & 6135 &3547 &760 &765 & 5072\\
twi  &--- & ---& ---& ---& ---     & ---    & ---     & ---    &2564 & 639 & 698 &3901 \\
vmw  &--- &--- &--- &--- &--- 1551  & 258  & 777   & 2586 &--- &--- &--- &--- \\
xho  &805 & 204&255 &1264 & ---     & 682  & 1594  & 2276 &2502 &559 &622 &3683 \\
yor  &8523 &2091 &4516 &15130 & 2992  & 497  & 1500  & 4989 &3336 &724 &819 &4879 \\
zul  &--- & ---& ---& --- & ---     & 875  & 2047  & 2922 &2940 &640 &728 &4308 \\
        \bottomrule
    \end{tabular}}
    \caption{Dataset distribution across different languages - AfriEmotion (train, dev, test, and total) and AfriSenti dataset. We adopt the same train-test-dev split from the data source papers: AfriEmo \cite{muhammad-etal-2025-semeval}, AfriSenti \cite{muhammad-etal-2023-afrisenti}, and AfriHate \cite{muhammad2025afrihate}.}
    \label{tab:dataset-split}
\end{table*}

\section{Baseline results from encoder-only LMs} \label{app:baseline}
We evaluate various encoder-only language models with the more difficult multi-label emotion classification task to select the best encoder-only model for continual learning. We found that AfroXLMR is better for the low-resourced African languages, and we continue our DAPT and TAPT training settings from this model. Table \ref{tab:baseline} shows baseline results for the AfriEmo dataset from different multilingual encoder-only models.
\begin{table*}[hbt!]
\centering
\resizebox{\textwidth}{!}{
\begin{tabular}{lllllllllllllll>{\columncolor{gray!20}}l>{\columncolor{gray!20}}l}
\toprule
\textbf{Model}&\texttt{afr}&\texttt{amh}&\texttt{ary}&\texttt{hau}&\texttt{ibo}&\texttt{kin}&\texttt{orm}&\texttt{pcm}&\texttt{ptMZ}&\texttt{som}&\texttt{swa}   &\texttt{tir}&\texttt{vmw}&\texttt{yor}&\texttt{xho}&\texttt{zul}\\
\midrule
\multicolumn{17}{l}{\textit{Baselines from general Multilingual models}}\\
LaBSE    &\textbf{37.76}&63.72&45.81&58.49&45.90&\textbf{50.64}&\textbf{43.30}&51.30
         &36.95&41.82&\textbf{27.53} &\textbf{47.23}&\textbf{21.13} &\textbf{32.55}&\textbf{31.39}&\textbf{18.16}\\
RemBERT  &37.14&\textbf{63.83}&\textbf{47.16}&\textbf{59.55}&\textbf{47.90}&46.29&12.63&
         \textbf{55.50}&\textbf{45.91}&\textbf{45.93}&22.65&46.28&12.14&9.22&12.73&15.26\\
mBERT    &26.87&26.69&36.87&47.33&37.23&35.61&39.84&48.42
         &14.81&31.13&22.99&25.16 &10.28&21.03&17.08&13.04\\
mDeBERTa &16.66&44.22&38.00&48.59&31.92&38.00&28.48&46.21
         &21.89&34.91&22.84&30.35 &11.13&17.88&22.86&13.87\\
\hline
\multicolumn{17}{l}{\textit{Baselines from African language-centric models}} \\
AfroLM      &21.60&54.78&30.35&57.31&42.46&38.97&41.84&47.12
            &17.81&32.43&20.08&38.22&15.98&24.31&13.67&10.72\\
AfriBERTa   &22.90&60.05&30.85&61.09&\textbf{49.05}&46.35&53.69&50.29&\textbf{23.15}
            &44.92&24.36&49.00&\textbf{20.29}&34.52&13.86&8.50\\
AfroXLMR    &43.66&68.97&47.62&64.30&26.27&52.39&52.28&55.39
            &22.09&48.78&30.74&57.22&21.18&28.65&13.52&6.90\\

\hline
\multicolumn{17}{l}{\textit{Continual pretraining from XLM-RoBERa-Large}} \\
XLMR-L      &38.69&54.99&38.31&52.99&38.72&35.06&26.67&53.77
            & 9.29&39.95& 6.62&18.58&13.53& 2.79&7.90&9.27\\
+ DAPT &25.94&60.08&44.41&59.81&42.91&44.61&43.61&53.74&30.57             
            &\textbf{41.69}&27.83&\textbf{49.55}&8.44&18.05&4.75&9.27\\
+ TAPT &\textbf{39.65}&\textbf{62.23}&\textbf{47.61}&\textbf{63.20}
            &\textbf{47.56}&39.09&45.25&\textbf{57.78}&\textbf{36.68}&38.68&\textbf{30.28}
            &44.74&\textbf{14.89}&\textbf{26.32}&\textbf{10.03}&\textbf{12.96}\\
+ DAPT +TAPT&23.00&60.81&42.91&60.44&43.44&\textbf{41.53}&\textbf{45.26}
            &53.90&29.83&40.05&24.02&48.28&9.32&18.53&6.19&4.63\\
\hline
\multicolumn{17}{l}{\textit{Continual Fine-tuning from  AfroXLMR-large }} \\
AfroXLMR &43.66&68.97&47.62&64.30&26.27&52.39&52.28&55.39&22.09&48.78&30.74&57.22&21.18&28.65&13.52&6.90\\
+ DAPT &44.57&\textbf{71.67}&\textbf{52.63}&\textbf{70.74}&\textbf{54.54}&\textbf{56.73}&\textbf{61.38}&\textbf{59.93}&36.80&\textbf{54.86}&\textbf{34.35}&\textbf{60.71}&\textbf{22.08}&\textbf{39.26}&8.54&6.72\\
+ TAPT &\textbf{49.61}&69.56&50.21&66.93&53.13&53.27&56.43&56.71&37.20&50.33&33.02&55.72&21.73&34.34&4.58&13.15\\    
+ DAPT + TAPT &44.17&71.04&51.63&69.77&54.26&54.47&58.75&58.89&\textbf{37.76}&52.65&32.74&57.12&19.80&35.89&21.30&13.81\\
\bottomrule
\end{tabular}
}
\caption{AfriEmo detail results using AfriSocial as DAPT and AfriSenti as TAPT. \textit{xho} and \textbf{zul} languages have no training set, and the results are in zero-shot. The results are from the average scores of five runs.}
\label{tab:baseline}
\end{table*}

\section{Training Details of DAFT and TAPT}\label{app:params}
\textbf{Hyper-parameters for adapters} We trained the task adapter using the following hyper-parameters: batch size of 32, 10 epochs, and learning rate of 5e-5. For TAPT, the parameters are similar to those of DAPT, except that the batch size is 8. We used their tokenizer, for XLMR - XLMR tokenizer, and AfroXLMR - AfroXLMR tokenizer. PyTorch was used for fine-tuning, and pre-trained models were sourced from Hugging Face. The domain-adaptive fine-tuning training is trained on three distributed GPUs for 3 days, whereas TAPT finishes in less than one hour. Following standard practice, we pass the final layer [CLS] token representation to a task-specific feedforward layer for prediction with three epochs. The reported results from fine-tuned pre-trained language models are the average results of five runs.

\section{Large Language Model Details} \label{app:llms-details} 
Multilingual Encoder-only, open-source, and proprietary model names and their sources are mentioned below. The results from LLMs are used from the work \cite{afrobench,muhammad-etal-2025-brighter}. All open-source LLMs are instruction-tuned versions. The various evaluation prompts are presented in the original works mentioned above. 
\subsection{Encoder-only Langauge Models}
\begin{itemize}
    \item LaBSE \cite{feng-etal-2022-language} - sentence-transformers/LaBSE 
    \item RemBERT \cite{chung2020rembert} - google/rembert 
    \item XLM-RoBERTa \cite{conneau-2020-unsupervised} - FacebookAI/xlm-roberta-base (large) 
    \item mDeBERTa \cite{he2021debertav3} - microsoft/mdeberta-v3-base 
    \item mBERT \cite{libovický2019lmbert}- google-bert\_bert-base-multilingual-cased 
    \item AfriBERTa \cite{ogueji-etal-2021-small} - castorini/afriberta\_large
    \item AfroXLMR \cite{adelani-etal-2024-sib} - Davlan/afro-xlmr-large-61L (76L)
    \item AfroLM \cite{dossou-etal-2022-afrolm}- bonadossou/afrolm\_active\_learning

    \end{itemize}

\subsection{Open-source LLMs}
\begin{itemize}
    \item Aya-101 13B \cite{aya-model} - CohereLabs/aya-101
    \item Llama 2 7B Chat \cite{touvron2023llama2} - meta-llama/Llama-2-7b-chat-hf
    \item Llama 3 8B \cite{grattafiori2024llama} - meta-llama/Meta-Llama-3-8B-Instruct
    \item Llama 3.1 (8B, 70B) \cite{grattafiori2024llama} - meta-llama/Llama-3.1-8B-Instruct and meta-llama/Llama-3.1-70B-Instruct, respectively.
    \item Gemma 1.1 7B \cite{team2024gemma} - google/gemma-1.1-7b-it
    \item Gemma 2 (9B, 27B) \cite{team2024gemma2} - google/gemma-2-2b-it and google/gemma-2-27b-it
    \item DeepSeek-R1-70 \cite{guo2025deepseek} - deepseek-ai/DeepSeek-R1-Distill-Llama-70B
    \item Mistral-8x7B \cite{jiang2024mixtral} - mistralai/Mixtral-8x7B-Instruct-v0.1
    \item Qwen2.5-72B \cite{yang2024qwen2} - Qwen/Qwen2.5-72B-Instruct
    \item Dolly-v2-12B  \cite{conover2023free} - databricks/dolly-v2-12b
\end{itemize}

\subsection{Propritary LLMs}
\begin{itemize}
    \item Gemini 1.5 Pro \cite{team2024gemini} - Gemini 1.5 Pro 002 accessed via Google API
    \item GPT-4o (Aug) \cite{openai2024gpt} -  the August 2024 version of the model is accessed through the OpenAI API
\end{itemize}

\newpage
\section{AfriSenti detail results}
Table \ref{tab:Afrisenti-all} shows all sentiment analysis results (AfriSenti dataset).
\begin{table}[htbp]
\centering
\resizebox{\textwidth}{!}{
\begin{tabular}{l*{14}{c}c}
\hline
\multirow{2}{*}{\textbf{Models}} & \multicolumn{14}{c}{\textbf{Languages}} & \multirow{2}{*}{\textbf{Avg.}} \\
\cline{2-15}
 & \texttt{amh} & \texttt{arq} & \texttt{ary} & \texttt{hau} & \texttt{ibo} & \texttt{kin} & \texttt{orm} & \texttt{pcm} & \texttt{por} & \texttt{swa} & \texttt{tir} & \texttt{tso} & \texttt{twi} & \texttt{yor} & \\
\hline
\multicolumn{16}{l}{\textit{Fine-tuned encoder-only models from the AfroXLMR baseline}} \\
AfroXLMR & 50.1 & 52.2 & 52.9 & 79.3 & 76.9 & 71.0 & 20.1 & 50.5 & 60.9 & 28.3 & 22.5 & 35.4 & 47.2 & 72.3 & 51.4 \\
AfroXLMR-Social & 57.2 & 64.6 & 62.3 & 62.7 & 79.8 & 72.7 & 24.3 & 52.1 & 64.8 & 61.4 & 24.5 & 38.8 & 56.0 & 74.6 & 56.9 \\
TAPT-Emo & 54.9 & 62.4 & 64.1 & 80.7 & 80.0 & 71.5 & 23.5 & 51.0 & 64.1 & 59.3 & 10.8 & 37.9 & 47.7 & 72.2 & \textbf{55.7} \\
TAPT-Hate & 54.5 & 59.4 & 52.8 & 80.1 & 78.3 & 69.7 & 43.0 & 50.3 & 62.8 & 57.3 & 16.2 & 36.2 & 50.2 & 70.9 & 55.8 \\
DAPT+TAPT-Emo & 55.8 & 63.4 & 63.1 & 82.7 & 80.4 & 69.5 & 28.9 & 52.0 & 63.8 & 54.9 & 28.9 & 36.7 & 54.5 & 73.7 & \textbf{57.7} \\
DAPT+TAPT-Hate & 56.3 & 59.7 & 62.1 & 82.0 & 80.2 & 70.1 & 23.6 & 51.2 & 62.1 & 58.5 & 21.9 & 40.0 & 55.8 & 73.4 & 56.9 \\
\hline
\multicolumn{16}{l}{\textit{Prompt-based proprietary models}} \\
LLaMAX3-8B & 55.2 & 55.5 & 51.0 & 61.7 & 54.6 & 53.2 & 33.6 & 56.0 & 41.3 & 54.1 & 43.5 & 48.0 & 39.0 & 50.4 & 49.8 \\
Llama-2-7B & 25.5 & 44.9 & 44.0 & 38.2 & 33.6 & 35.4 & 24.7 & 60.8 & 31.2 & 33.8 & 33.5 & 46.1 & 48.9 & 43.7 & 38.9 \\
Llama-3.1-70B & 40.0 & 47.5 & 53.5 & 52.6 & 52.2 & 48.5 & 41.4 & 52.6 & 35.9 & 61.5 & 28.2 & 43.3 & 45.8 & 54.3 & 47.0 \\
Llama-3.1-8B & 66.4 & 57.1 & 51.9 & 55.4 & 50.1 & 48.7 & 35.9 & 64.2 & 33.6 & 54.3 & 49.8 & 48.8 & 42.3 & 50.9 & 50.7 \\
Llama-3-8B & 46.3 & 51.0 & 46.1 & 38.5 & 36.1 & 38.4 & 28.2 & 60.2 & 27.9 & 37.8 & 38.0 & 43.3 & 47.7 & 45.1 & 41.8 \\
Aya-101 & 76.8 & 67.8 & 58.1 & 61.2 & 47.5 & 61.1 & 37.4 & 70.1 & 48.8 & 47.5 & 71.2 & 50.8 & 44.7 & 57.0 & 57.1 \\
Gemma-1.1-7B & 24.4 & 43.1 & 42.0 & 37.9 & 34.7 & 32.0 & 25.9 & 66.5 & 37.4 & 37.0 & 32.4 & 50.0 & 48.7 & 43.8 & 39.7 \\
Gemma-2-27B & 70.7 & 65.8 & 59.0 & 64.8 & 60.4 & 59.1 & 37.3 & 76.0 & 42.8 & 55.6 & 58.9 & 50.0 & 54.3 & 65.5 & \textbf{58.6} \\
Gemma-2-9B & 70.1 & 62.0 & 56.4 & 61.4 & 58.2 & 56.1 & 37.9 & 66.8 & 46.6 & 58.7 & 55.4 & 43.7 & 48.1 & 55.4 & 55.5 \\
\hline
\multicolumn{16}{l}{\textit{Prompt-based proprietary models}} \\
Gemini-1.5 & 77.5 & 70.9 & 63.7 & 70.1 & 56.9 & 68.3 & 42.8 & 74.5 & 46.4 & 55.2 & 70.2 & 55.9 & 49.3 & 74.3 & \textbf{62.6} \\
GPT-4o & 75.6 & 72.3 & 61.2 & 68.6 & 67.8 & 71.6 & 43.1 & 67.1 & 62.1 & 57.9 & 61.5 & 46.5 & 51.3 & 70.2 & \textbf{62.6} \\
\hline
\end{tabular}
}
\caption{AfriSenti Model Performance Across Various Languages}
\label{tab:Afrisenti-all}
\end{table}

\section{AfriHate results}
Table \ref{tab:afrihate-all} shows all hate speech classification results (AfriHate dataset).
\begin{table}[htbp]
\centering
\resizebox{\textwidth}{!}{
\begin{tabular}{l*{15}{c}c}
\hline
\multirow{2}{*}{\textbf{Models}} & \multicolumn{15}{c}{\textbf{Languages}} & \multirow{2}{*}{\textbf{Avg.}} \\
\cline{2-16}
 & \texttt{amh} & \texttt{arq} & \texttt{ary} & \texttt{hau} & \texttt{ibo} & \texttt{kin} & \texttt{orm} & \texttt{pcm} & \texttt{som} & \texttt{swa} & \texttt{tir} & \texttt{twi} & \texttt{xho} & \texttt{yor} & \texttt{zul} & \\
\hline
\multicolumn{17}{l}{\textit{Fine-tuned encoder-only models from the AfroXLMR baseline}} \\
AfroXLMR & 73.5 & 43.4 & 75.1 & 81.6 & 82.8 & 75.3 & 67.2 & 64.9 & 55.7 & 91.5 & 50.2 & 46.9 & 50.9 & 53.4 & 54.5 & 64.5 \\
DAPT & 78.6 & 46.0 & 75.6 & 80.8 & 88.1 & 78.8 & 74.1 & 67.6 & 55.6 & 91.2 & 55.9 & 48.4 & 59.2 & 77.9 & 55.4 & \textbf{68.9} \\
TAPT-Emo & 73.1 & 46.2 & 75.1 & 77.8 & 87.7 & 77.5 & 69.0 & 67.7 & 57.3 & 91.9 & 54.2 & 49.3 & 55.1 & 53.7 & 56.0 & 66.1 \\
TAPT-Senti & 73.3 & 43.8 & 71.7 & 81.9 & 86.5 & 76.3 & 67.7 & 66.9 & 55.0 & 91.3 & 57.0 & 48.9 & 51.3 & 54.5 & 55.0 & 65.4 \\
DAPT + TAPT-Emo & 78.1 & 44.2 & 77.5 & 82.1 & 87.7 & 78.4 & 71.6 & 69.9 & 56.8 & 91.2 & 32.7 & 49.0 & 58.7 & 54.8 & 55.5 & 65.9 \\
DAPT + TAPT-Senti & 77.6 & 43.8 & 76.1 & 81.6 & 79.4 & 77.9 & 72.0 & 67.5 & 53.7 & 91.5 & 41.2 & 48.2 & 54.6 & 54.8 & 55.8 & 65.0 \\
\hline
\multicolumn{17}{l}{\textit{Prompt-based proprietary models}} \\
Gemma-1.1-7B & 23.0 & 27.4 & 24.5 & 26.0 & 16.7 & 29.9 & 27.9 & 30.2 & 27.2 & 27.4 & 17.3 & 14.2 & 23.3 & 25.0 & 22.5 & 24.3 \\
Llama-2-7B & 14.5 & 22.4 & 22.2 & 24.4 & 20.2 & 22.4 & 31.3 & 9.4 & 27.1 & 24.8 & 11.7 & 15.8 & 24.8 & 23.1 & 26.8 & 21.9 \\
Llama-3-8B & 26.5 & 31.8 & 28.5 & 24.5 & 19.7 & 36.5 & 37.1 & 38.8 & 17.8 & 34.3 & 28.4 & 14.4 & 25.0 & 25.9 & 28.4 & 27.9 \\
LLaMAX3-8B& 37.2 & 33.6 & 31.5 & 30.7 & 19.4 & 38.2 & 38.2 & 34.4 & 27.6 & 28.9 & 27.4 & 13.9 & 23.7 & 24.4 & 29.0 & 28.6 \\
Llama-3.1-8B & 23.3 & 30.7 & 22.9 & 25.4 & 13.9 & 31.9 & 35.7 & 24.9 & 26.7 & 21.7 & 21.9 & 9.9 & 22.4 & 19.4 & 23.3 & 23.6 \\
gemma-2-9b & 33.2 & 33.8 & 33.2 & 24.1 & 25.1 & 33.6 & 26.7 & 54.9 & 13.6 & 46.4 & 26.8 & 29.1 & 20.0 & 30.5 & 20.1 & 29.9 \\
Aya-101-13B & 31.3 & 32.1 & 28.9 & 33.3 & 22.1 & 32.8 & 26.8 & 37.8 & 35.8 & 41.3 & 29.6 & 13.8 & 28.7 & 26.8 & 29.8 & 30.0 \\
Gemma-2-27B & 48.4 & 49.1 & 53.8 & 34.8 & 42.8 & 52.7 & 39.8 & 60.9 & 39.6 & 70.9 & 35.4 & 38.1 & 30.6 & 54.0 & 35.0 & 45.5 \\
Llama-3.1-70B & 53.0 & 57.0 & 60.6 & 41.2 & 48.4 & 50.9 & 44.6 & 62.4 & 39.8 & 67.0 & 41.0 & 37.9 & 32.7 & 56.2 & 46.3 & \textbf{49.0} \\
\hline
\multicolumn{16}{l}{\textit{Prompt-based proprietary models}} \\
Gemini-1.5 pro & 56.1 & 70.6 & 68.2 & 61.4 & 66.9 & 64.2 & 57.6 & 65.0 & 60.8 & 80.5 & 37.5 & 50.6 & 58.0 & 73.1 & 55.4 & 62.1 \\
GPT-4o & 56.0 & 69.7 & 75.5 & 59.2 & 69.7 & 60.1 & 53.5 & 65.2 & 68.5 & 78.0 & 42.4 & 51.2 & 63.7 & 74.5 & 58.7 & \textbf{63.5} \\
\hline
\end{tabular}
}
\caption{AfriHate Model Performance Across Various Languages}
\label{tab:afrihate-all}
\end{table}

\newpage
\section{AfriEmo detail results}
Table \ref{tab:afriemo-all} shows all fine-grained multi-label emotion classification results (AfriEmo dataset).

\begin{table}[htbp]
\centering
\resizebox{\textwidth}{!}{
\begin{tabular}{l*{17}{c}c}
\hline
\multirow{2}{*}{\textbf{Models}} & \multicolumn{17}{c}{\textbf{Languages}} & \multirow{2}{*}{\textbf{Avg.}} \\
\cline{2-18}
 & \texttt{afr} & \texttt{amh} & \texttt{arq} & \texttt{ary} & \texttt{hau} & \texttt{ibo} & \texttt{kin} & \texttt{orm} & \texttt{pcm} & \texttt{ptMZ} & \texttt{som} & \texttt{swa} & \texttt{tir} & \texttt{vmw} & \texttt{yor} & \texttt{xho} & \texttt{zul} & \\
\hline
\multicolumn{19}{l}{\textit{Fine-tuned encoder-only models from the AfroXLMR baseline and others}} \\
AfroXLMR & 43.7 & 69.0 & 44.9 & 47.6 & 64.3 & 26.3 & 52.4 & 52.3 & 55.4 & 22.1 & 48.8 & 30.7 & 57.2 & 21.2 & 28.7 & 13.5 & 6.9 & 40.3 \\
+ DAPT & 44.6 & 71.7 & 51.3 & 52.6 & 70.7 & 54.5 & 56.7 & 61.4 & 59.9 & 36.8 & 54.9 & 34.4 & 60.7 & 22.1 & 39.9 & 8.5 & 6.7 & \textbf{46.3} \\
+ TAPT-Senti & 49.6 & 69.6 & 49.0 & 50.2 & 66.9 & 53.1 & 53.3 & 56.4 & 56.7 & 37.2 & 50.3 & 33.0 & 55.7 & 21.7 & 34.3 & 14.6 & 13.2 & 45.0 \\
+ TAPT-Hate & 47.2 & 67.2 & 48.4 & 48.8 & 61.2 & 51.3 & 53.5 & 52.2 & 56.6 & 31.0 & 49.6 & 31.9 & 55.8 & 19.3 & 32.9 & 11.0 & 9.9 & 42.8 \\
+ DAPT + TAPT-Senti & 44.2 & 71.0 & 48.7 & 51.6 & 69.8 & 54.3 & 54.5 & 58.8 & 58.9 & 37.8 & 52.7 & 32.7 & 57.1 & 19.8 & 35.9 & 21.3 & 13.8 & 46.1 \\
+ DAPT + TAPT-hate & 46.2 & 70.8 & 46.6 & 48.8 & 70.0 & 54.2 & 53.5 & 56.8 & 58.2 & 34.8 & 52.7 & 31.4 & 57.1 & 19.7 & 33.0 & 9.4 & 5.0 & 44.0 \\
LaBSE & 35.1 & 63.7 & 35.9 & 42.8 & 38.5 & 18.1 & 30.4 & 43.3 & 33.3 & 31.4 & 41.8 & 21.7 & 47.2 & 9.7 & 11.6 & 31.4 & 18.2 & 32.6 \\
RemBERT & 35.0 & 63.8 & 33.8 & 35.5 & 32.0 & 7.5 & 18.4 & 12.6 & 1.0 & 29.7 & 45.9 & 19.0 & 46.3 & 5.2 & 5.3 & 12.7 & 15.3 & 24.7 \\
XLM-R & 41.7 & 46.9 & 35.9 & 33.9 & 16.7 & 10.4 & 13.1 & 17.9 & 21.1 & 7.3 & 25.4 & 16.9 & 35.9 & 12.7 & 6.6 & 11.5 & 10.9 & 21.5 \\
mBERT & 17.0 & 26.7 & 31.4 & 24.8 & 15.6 & 9.9 & 20.9 & 39.8 & 22.6 & 13.5 & 31.1 & 18.6 & 25.2 & 12.1 & 9.6 & 17.1 & 13.0 & 20.5 \\
mDeBERTa & 33.3 & 44.2 & 35.9 & 36.3 & 32.8 & 9.5 & 17.3 & 28.5 & 25.4 & 24.5 & 34.9 & 14.9 & 30.4 & 11.7 & 10.0 & 22.9 & 13.9 & 25.1 \\
\hline
\multicolumn{19}{l}{\textit{Prompt-based proprietary models}} \\

Qwen2.5-72B & 60.2 & 37.8 & 37.8 & 52.8 & 43.8 & 37.4 & 32.0 & 31.6 & 38.7 & 40.4 & 28.6 & 27.4 & 31.1 & 20.4 & 25.0 & 29.6 & 22.0 & 35.1 \\
Dolly-v2-12B & 23.6 & 5.1 & 38.6 & 24.3 & 29.4 & 24.3 & 19.7 & 22.9 & 34.4 & 16.7 & 19.8 & 17.6 & 1.5 & 16.0 & 16.0 & 24.1 & 14.7 & 20.5 \\
Llama-3.3-70B & 61.3 & 42.8 & 55.8 & 45.0 & 50.9 & 33.2 & 34.4 & 29.8 & 48.7 & 34.1 & 32.5 & 29.5 & 32.9 & 19.0 & 23.7 & 30.8 & 21.5 & \textbf{36.8} \\
Mixtral-8x7B & 53.7 & 29.0 & 45.3 & 35.1 & 40.4 & 31.9 & 26.4 & 24.3 & 45.6 & 36.5 & 25.6 & 26.5 & 27.2 & 19.0 & 19.7 & 22.9 & 20.4 & 31.1 \\
DeepSeek-R1-70B & 43.7 & 36.9 & 50.9 & 47.2 & 51.9 & 32.9 & 32.5 & 28.2 & 45.0 & 39.6 & 26.6 & 33.3 & 26.5 & 19.1 & 27.4 & 29.1 & 20.4 & 34.8 \\
\hline
\end{tabular}
}
\caption{AfriEmo Model Performance Across Various Languages}
\label{tab:afriemo-all}
\end{table}

\section{AfriSocial Data Sources}\label{app:source}

\paragraph{ X (Twitter) Source} There is an X domain corpus and model for high-resource languages such as XLM-T \cite{XLMR-T:2022:LREC} to evaluate and improve task datasets sourced from X. However, there is a scarcity of corpora specializing in the social domain for low-resource African languages. The tweets are scraped over a different time until June 2023, before the change of an X policy that restricts their data for academic research.

\paragraph{ News Source} News platforms are the most common data source for African languages. Companies also stream their news on the X platform. While more formal, news websites also provide a platform for public discourse, comments, and reactions, often including opinion pieces and user-generated comments. The source news websites are British Broadcasting Corporation  (BBC) news\footnote{\url{https://www.bbc.com/}}, Akan news\footnote{\url{https://akannews.com}}, Global Voice News\footnote {\url{https://mg.globalvoices.org/}}, isolezwelesixhosa\footnote{\url{https://www.isolezwelesixhosa.co.za/}}, isolezwe\footnote{\url{https://www.isolezwe.co.za/}}, and others. 

\end{document}